\documentclass[utf8]{FrontiersinHarvard}
\usepackage{url,hyperref,lineno,microtype,subcaption}
\usepackage[onehalfspacing]{setspace}
\usepackage{etoolbox}

\def\keyFont{\fontsize{8}{11}\helveticabold }
\def\firstAuthorLast{Chen {et~al.}} 
\def\Authors{Yang Chen\,$^{1,*}$, Sanglin Zhao\,$^{2}$, Baoyu Chen\,$^{3}$ and M{\aa}ns Gustaf\,$^{4,*}$} 

\def\Address{$^{1}$School of Mathematics and Statistics, Xiamen University of Technology, Xiamen, China (email: 2019000009@xmut.edu.cn) \\
$^{2}$Hunan University of Finance and Economics, Changsha, China (email: 202205640108@mails.hufe.edu.cn) \\
$^{3}$Dongguan University of Technology, Dongguan, China (email: 2022463030801@dgut.edu.cn) \\
$^{4}$School of Business, Society and Engineering, M\"{a}lardalens University, V\"{a}ster\r{a}s, Sweden (gustafedu@yeah.net) \\
$^{*}$Correspondence: 2019000009@xmut.edu.cn (Y.C.); gustafedu@yeah.net (M.G.)} 

\def\corrAuthor{Yang Chen and M{\aa}ns Gustaf} 
\def\corrEmail{2019000009@xmut.edu.cn; gustafedu@yeah.net} 

\makeatletter
\def\@correspondance{}
\providecommand{\@extraAuth}{} 
\makeatother

\begin{document}
\onecolumn
\firstpage{1}

\title[Adaptive Contrast for Fetal US]{Adaptive Contrast Adjustment Module: A Clinically-Inspired Plug-and-Play Approach for Enhanced Fetal Plane Classification}

\author[\firstAuthorLast ]{\Authors}
\address{}

\maketitle

\begin{abstract}
Fetal ultrasound standard plane classification is essential for reliable prenatal diagnosis but faces inherent challenges, including low tissue contrast, boundary ambiguity, and operator-dependent image quality variations. To overcome these limitations, we propose a plug-and-play adaptive contrast adjustment module (ACAM), whose core design is inspired by the clinical practice of doctors adjusting image contrast to obtain clearer and more discriminative structural information. The module employs a shallow texture-sensitive network to predict clinically plausible contrast parameters, transforms input images into multiple contrast-enhanced views through differentiable mapping, and fuses them within downstream classifiers. Validated on a multi-center dataset of 12,400 images across six anatomical categories, the module consistently improves performance across diverse models, with accuracy of lightweight models increasing by 2.02 percent, accuracy of traditional models increasing by 1.29 percent, and accuracy of state-of-the-art models increasing by 1.15 percent. The innovation of the module lies in its content-aware adaptation capability, replacing random preprocessing with physics-informed transformations that align with sonographer workflows while improving robustness to imaging heterogeneity through multi-view fusion. This approach effectively bridges low-level image features with high-level semantics, establishing a new paradigm for medical image analysis under real-world image quality variations.

\keyFont{ \section{Keywords:} Fetal ultrasound, Clinically-inspired module, Adaptive contrast adjustment, Robust medical image analysis}
\end{abstract}

\section{Introduction}

Ultrasound possesses advantages such as safety, convenience, non-invasiveness, and absence of radiation, making it widely used in critical areas such as prenatal fetal screening \cite{maher2024fetal, wittek2025innovations, al2024radiological, miller2020diagnostic,wang2018image}. Obtaining standardized fetal ultrasound planes is crucial for improving diagnostic accuracy and reducing the likelihood of missed severe fetal abnormalities. However, this process faces multiple challenges: operators must have a thorough understanding of fetal anatomy, and clinical experience or equipment conditions may be insufficient. In addition, the increasing complexity of screening environments, the growing demand for fetal examinations, and the shortage of trained ultrasound professionals further exacerbate the difficulty of performing high-quality manual screening. Under these circumstances, there is an urgent need to develop automated recognition technologies to assist sonographers in accurately and efficiently identifying standard fetal trunk planes. Such technologies can reduce the risk of missed diagnoses, enhance screening efficiency, and provide more reliable and safe technical support for prenatal diagnosis.

Deep learning techniques, owing to their powerful capabilities, have been widely applied across various medical fields \cite{cai2024developing, zhu2024covidllm, ou2024rtseg,mykula2024diffusion}. Recent studies have increasingly focused on algorithms for fetal ultrasound plane analysis \cite{zhu2025singular, zhu2025multi, boumeridja2025enhancing, montero2021generative, yousefpour2023enhancing}. However, fetal plane images often contain complex anatomical structures, ambiguous boundaries, or low-contrast regions, making it difficult to effectively highlight certain critical details and thereby limiting the discriminative capability of models. To address these challenges, we propose an Adaptive Contrast Adjustment Module (ACAM) that enables the model to automatically adjust image contrast based on content, generating multiple contrast-enhanced views and fusing their information. This approach enhances the representation of texture details and improves the model’s classification performance on complex fetal plane images. Our design is inspired by the practical procedures that clinicians follow when selecting fetal planes during ultrasound examination.

In clinical practice, sonographers often adjust image contrast to highlight key anatomical structures, thereby obtaining clearer and more discriminative ultrasound images \cite{smith1982contrast,mehta2017vascular}. Inspired by this, we introduce an adaptive contrast adjustment mechanism into our model. First, the input image is processed by a decision network that predicts K potential contrast parameters. These parameters are then mapped to a predefined fixed range through a differentiable function to ensure numerical stability and maintain model trainability. Based on these contrast parameters, the input image is transformed to generate K contrast-enhanced versions, effectively introducing multiple perspectives or styles into the training data. These images are subsequently fed into a convolutional neural network for feature extraction and classification. Since contrast adjustment primarily depends on local image details rather than high-level semantic features, we employ a shallow convolutional network as the decision module to extract local texture information and generate the contrast parameters. The shallow network captures fine-grained details such as textures and edges. This approach offers interpretability and generalization potential: the contrast parameters directly control image brightness and contrast, making the transformation easily visualizable and more intuitive compared to black-box manipulations of high-level features. Furthermore, by explicitly simulating multiple contrast scenarios, the model learns representations that are more robust to variations in illumination and contrast, thereby improving generalization across different acquisition conditions or domains.

In addition, our module is plug-and-play and only integrated at the lower layers of the model, facilitating implementation. We integrated our module into traditional robust models, lightweight models, and state-of-the-art models, conducting experiments on eight different architectures. The results demonstrate that integrating our module consistently improves model performance. The advantages of our module can be summarized as follows:

\begin{itemize}
    \item Our module simulates the process by which clinicians adjust image contrast. It has the ability to adaptively modify image contrast, generating multiple images with different contrast parameters and fusing them to produce images with clearer textures. This process effectively enhances the model’s sensitivity to fine-grained details and improves its robustness.
    
    \item In our proposed model, the input image is first processed by a shallow convolutional network to extract texture features. Based on these features, the network predicts multiple candidate contrast values, which are then used to perform contrast enhancement on the image, further improving the representation of features.
    
    \item We integrated this module into lightweight convolutional neural networks, traditional robust convolutional models, and advanced architectures, and conducted extensive empirical evaluations. Through comparative experiments, ablation studies, and heatmap visualizations, we demonstrated the effectiveness and generalizability of the proposed module in enhancing model performance.
    
\end{itemize}

\section{Method}
\subsection{Linear Contrast }

Image contrast enhancement can be achieved through either linear or nonlinear gray-level transformations, with the basic goal of stretching or compressing the distribution range of pixel intensities, thereby emphasizing the intensity differences across regions of the image. Let the original grayscale image be denoted as $I(x,y)$, where $(x,y)$ represents the pixel location in the image. A commonly used linear contrast adjustment method can be formulated as:

\begin{equation}
    I'(x,y) = \alpha \cdot (I(x,y) - \mu) + \mu,
\end{equation}

where $I'(x,y)$ represents the adjusted pixel intensity, $\alpha > 0$ is the contrast scaling factor (typically referred to as the contrast gain), and $\mu$ is the mean intensity (brightness center) of the image, defined as:

\begin{equation}
    \mu = \frac{1}{HW} \sum_{x=1}^{H} \sum_{y=1}^{W} I(x,y),
\end{equation}

with $H$ and $W$ denoting the image height and width, respectively.  

When $\alpha > 1$, the contrast of the image is increased, whereas $\alpha < 1$ reduces the contrast.

\subsection{The mechanism of the ACAM module}
\begin{figure}
    \centering
    \includegraphics[width=0.9\linewidth]{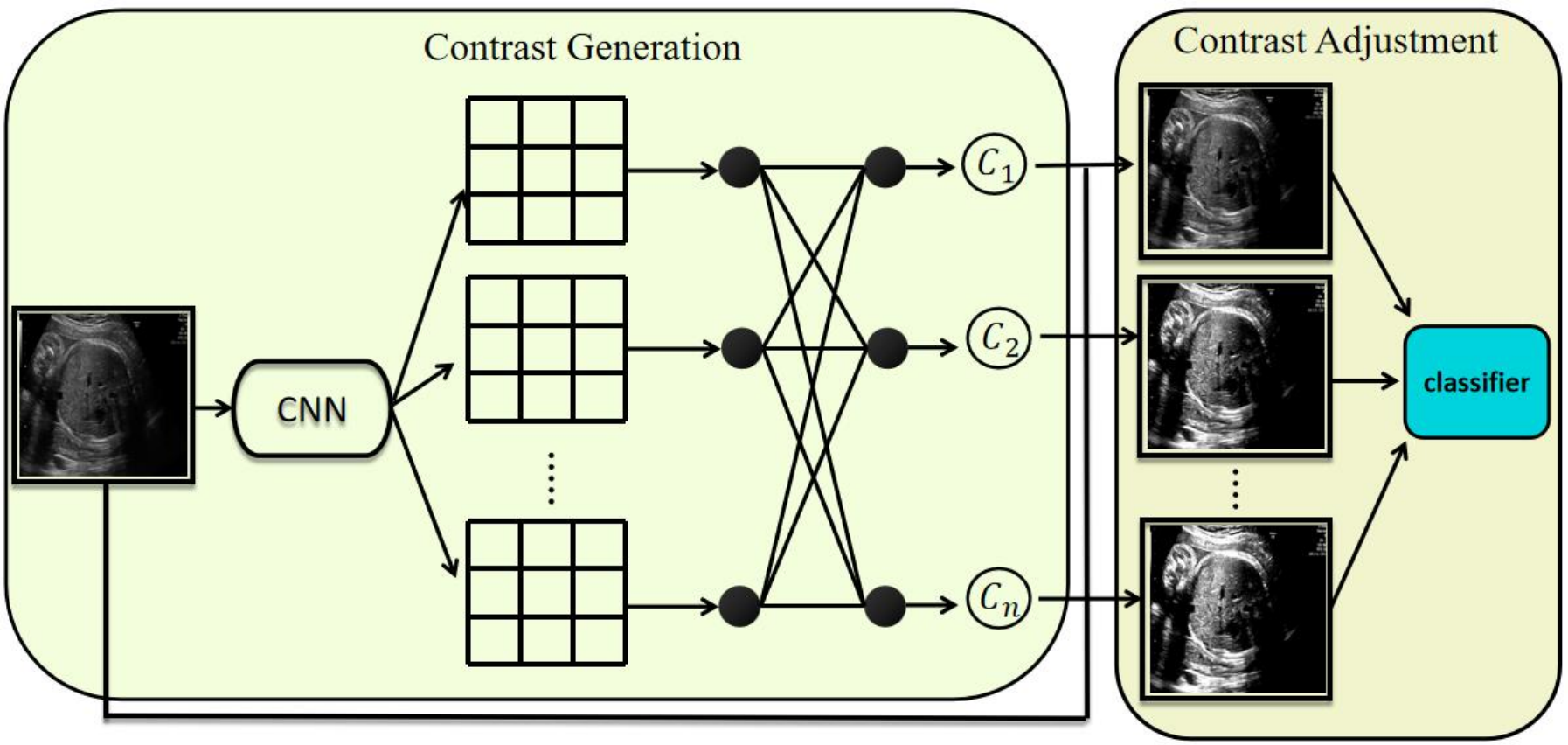}
    \caption{Architecture of the proposed module. It consists of two components: contrast generation and contrast adjustment. The contrast generation component predicts k distinct contrast parameters, which are subsequently used by the contrast adjustment component to transform the input image accordingly.}
    \label{module}
\end{figure}

The architecture of our proposed module is illustrated in Figure \ref{module}. The input is a grayscale image of dimensions [1, H, W]. The first stage of the module generates a set of contrast values from the input image to guide subsequent operations. We hypothesize that contrast prediction depends more heavily on local textural details than on high-level semantic features. To this end, the module adopts a shallow network consisting of convolutional layers, a global average pooling layer, and fully connected layers. These predicted contrast values are then utilized as adaptive parameters within the model.

\begin{equation}
    I_{kij} = c_k \left( I_{kij} - \sum_{ij} I_{kij} \right) + \sum_{ij} I_{kij}
\end{equation}

 These images constitute the output of our module and serve as input to subsequent decision-making models such as MedMamba.

\subsection{Implementation Details}
This study is based on a large-scale prenatal screening ultrasound image dataset \cite{burgos2020evaluation}, which was collected from two hospitals and encompasses multiple operators as well as different ultrasound device models. All images were manually annotated by a single obstetrics expert and categorized into six classes: four commonly used fetal standard planes (abdominal, brain, femur, and thoracic), the maternal cervix plane for preterm screening, and a general class including other less common planes. The final dataset comprises over 12,400 images from 1,792 patients, and it was split into training and test sets at a ratio of 7:3. All experiments were conducted using Python 3.9 and the PyTorch 2.0.1+cu117 framework, on a system equipped with an Intel i7-12650H processor and an NVIDIA RTX2080Ti GPU. Detailed settings of the model parameters and baseline models are provided in Table \ref{tab1}, with the number of generated contrast images n set to 10.

\renewcommand{\arraystretch}{1.5}

\begin{table}[!ht]
    \centering
    \caption{Hyperparameter settings used during model training}
    \begin{tabular}{p{5cm}p{7cm}}
    \hline
        Hyperparameter & Value  \\ \hline
        Batch size & 64  \\ 
        Epoch & 20  \\ 
        Lr & 0.001  \\ 
        optimizer & Adam (\cite{bock2019proof})  \\ 
        Loss Function & Crossentropy (\cite{mao2023cross}) \\ \hline
    \end{tabular}
    \label{tab1}
\end{table}

\section{Results}
\subsection{Evaluation Metrics and Baseline Models}

In this study, multiple widely adopted evaluation metrics are employed to systematically analyze model performance. Accuracy (ACC) reflects the overall correctness of predictions; however, it may be misleading in scenarios with imbalanced class distributions. Recall measures the model's ability to correctly identify positive samples, which is particularly crucial in medical image analysis, as higher recall helps reduce the risk of missed diagnoses. Precision evaluates the proportion of predicted positive samples that are truly positive, thereby reducing the likelihood of false alarms. The F1-score, defined as the harmonic mean of precision and recall, provides a balanced assessment of both metrics. 

Here, we denote the standard confusion matrix terms as follows:  
\begin{itemize}
    \item $TP$ (True Positive): number of correctly predicted positive samples,
    \item $TN$ (True Negative): number of correctly predicted negative samples,
    \item $FP$ (False Positive): number of negative samples incorrectly predicted as positive,
    \item $FN$ (False Negative): number of positive samples incorrectly predicted as negative.
\end{itemize}

Based on these definitions, the metrics are computed as:

\begin{equation}
\text{Accuracy} = \frac{TP + TN}{TP + TN + FP + FN}
\end{equation}

\begin{equation}
\text{Precision} = \frac{TP}{TP + FP}
\end{equation}

\begin{equation}
\text{Recall} = \frac{TP}{TP + FN}
\end{equation}

\begin{equation}
F1 = \frac{2 \times \text{Precision} \times \text{Recall}}{\text{Precision} + \text{Recall}}
\end{equation}

In addition, to comprehensively characterize the model's classification capability across different decision thresholds, we introduce the Receiver Operating Characteristic (ROC) curve and employ the Area Under the Curve (AUC) as a performance indicator. Similarly, the Precision–Recall (PR) curve is utilized to illustrate prediction accuracy at varying recall levels, with the Average Precision (AP) computed to intuitively reflect the model's ability in target detection tasks. 

To evaluate the effectiveness of the proposed model, we compare it against several established deep learning architectures, including EfficientNet \cite{kashyap2023review}, InceptionV3 \cite{szegedy2016rethinking}, VGG \cite{gunasekaran2024disease}, ResNet \cite{xu2023resnet}, MobileNet \cite{han2022triple}, ShuffleNet \cite{hou2025fault}, ConvNeXt\cite{sangeetha2024survey}, and MedMamba\cite{bansal2024comprehensive}.

\subsection{Comparison experiment}
The performance comparison of the models is presented in Table \ref{tab2}. Experimental results indicate that, regardless of whether they are lightweight models, traditional robust architectures, or state-of-the-art deep learning models, all achieved strong performance on the test set, with overall accuracy exceeding the 90\% baseline. Specifically, EfficientNet and InceptionV3, as representatives of classic architectures, achieved top-1 accuracies of 92.1\% and 92.3\%, respectively. Notably, our proposed ACAM-MedMamba model demonstrated the best performance, achieving an accuracy and F1-score of 93.47\%, significantly outperforming the other comparative models. Moreover, when ACAM was integrated into other backbone networks, the models similarly exhibited enhanced performance, confirming the generalization capability of the proposed architecture across different backbones.

\begin{table}[!ht]
    \centering
    \caption{Ablation study results of our module integrated into different models, as well as comparisons with other models. The best-performing values are highlighted in bold.}
    \begin{tabular}{p{5cm}p{2cm}p{2cm}p{2cm}p{2cm}}
    \hline
        Model & ACC & Precision & Recall & F1-score  \\ \hline
        ACAM-ResNet & 0.9301 & 0.9318 & 0.9301 & 0.9300  \\ 
        ResNet & 0.9172 & 0.9203 & 0.9172 & 0.9167  \\ \hline
        ACAM-MedMamba & 0.9347 & 0.9351 & 0.9347 & 0.9347  \\ 
        MedMamba & 0.9232 & 0.9266  & 0.9232 & 0.9236   \\ \hline
        ACAM-ShuffleNet & 0.9130 & 0.9125 & 0.9130 & 0.9125  \\ 
        ShuffleNet & 0.8928 & 0.8920 & 0.8928 & 0.8894  \\ \hline
        EfficientNet & 0.9226 & 0.9237 & 0.9226 & 0.9226  \\ 
        InceptionV3 & 0.9232 & 0.9240 & 0.9232 & 0.9216  \\ 
        MobileNet & 0.9027 & 0.9076 & 0.9027 & 0.9036  \\ 
        VGG & 0.9073 & 0.9100 & 0.9073 & 0.9075  \\ 
        ConVNeXt & 0.8923 & 0.8951 & 0.8923 & 0.8899 \\ \hline
    \end{tabular}
    \label{tab2}
\end{table}

\subsection{Ablation study}
The results of the ablation study are summarized in Table \ref{tab2}. It can be seen that, regardless of whether the backbone is a traditional model (ResNet), a lightweight model (ShuffleNet), or a state-of-the-art model (MedMamba), integrating the proposed module leads to a significant performance improvement, with an average gain of 1.48\%. This consistent enhancement across different architectures demonstrates the effectiveness and generality of the proposed module.

\begin{figure}
    \centering
    \includegraphics[width=0.9\linewidth]{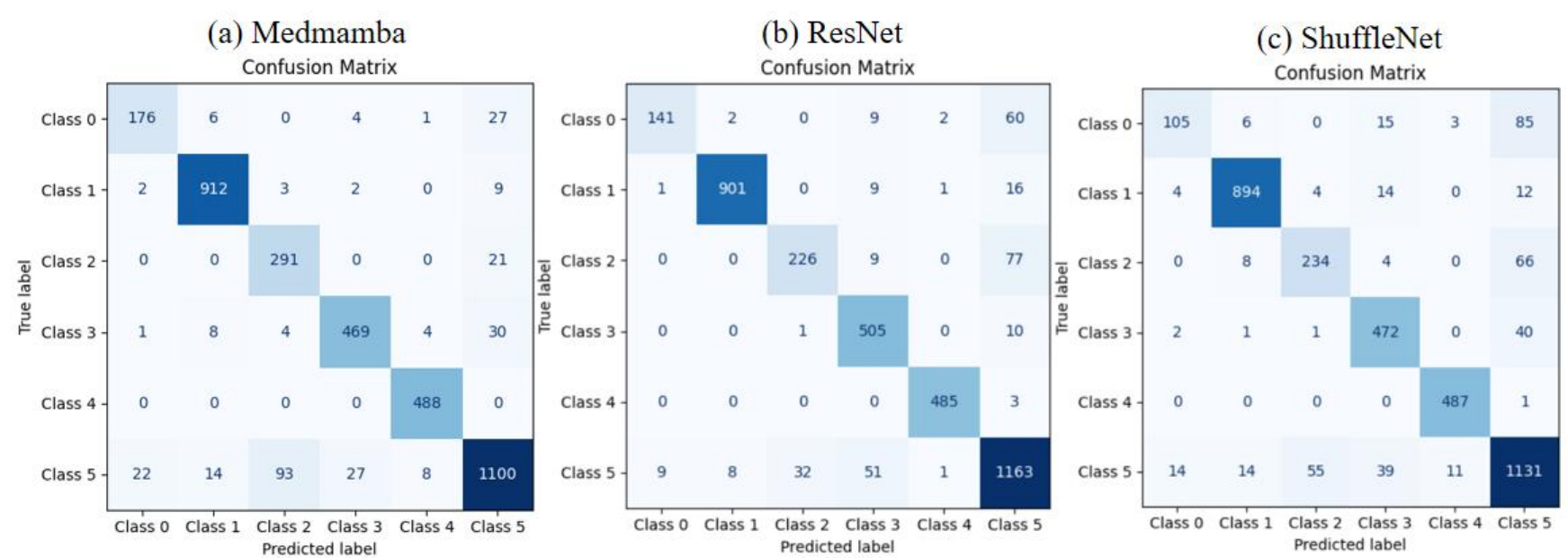}
    \includegraphics[width=0.9\linewidth]{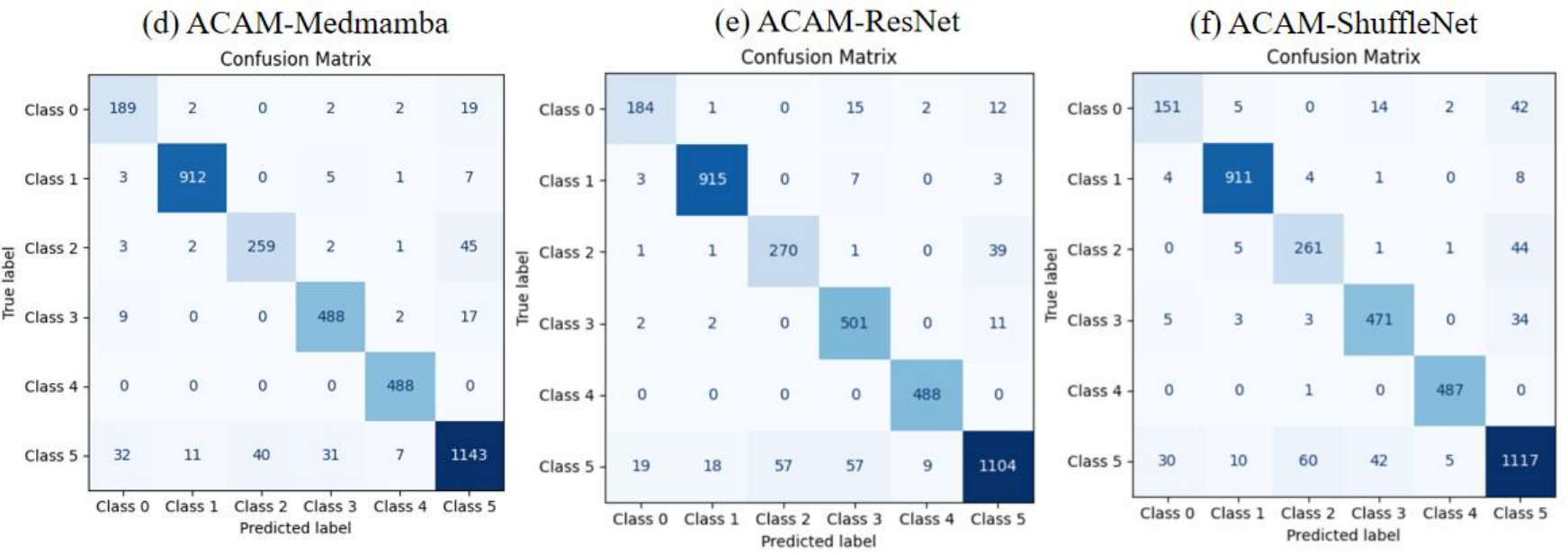}
    \caption{Comparison of confusion matrices for three models before and after integrating the proposed module. (a)–(c) show the classification performance of the baseline models, while (d)–(f) illustrate the improvements achieved after incorporating the module.}
    \label{conf}
\end{figure}

A comparison of confusion matrices, as shown in Figure \ref{conf}, reveals that the ACAM module consistently improves classification performance across lightweight models (ShuffleNet), traditional models (ResNet), and state-of-the-art models (MedMamba). In particular, the classification accuracy for classes 0 and 1 is significantly enhanced in all models, with a substantial reduction in misclassifications. For class 5, most cases also show improved precision after module integration. These results highlight that ACAM can robustly optimize feature discrimination for both common and challenging classes across various backbone networks. Furthermore, the module effectively mitigates inter-class confusion, especially in models prone to overfitting or with limited representational capacity, confirming its generalization and robustness.

\begin{figure}
    \centering
    \includegraphics[width=0.9\linewidth]{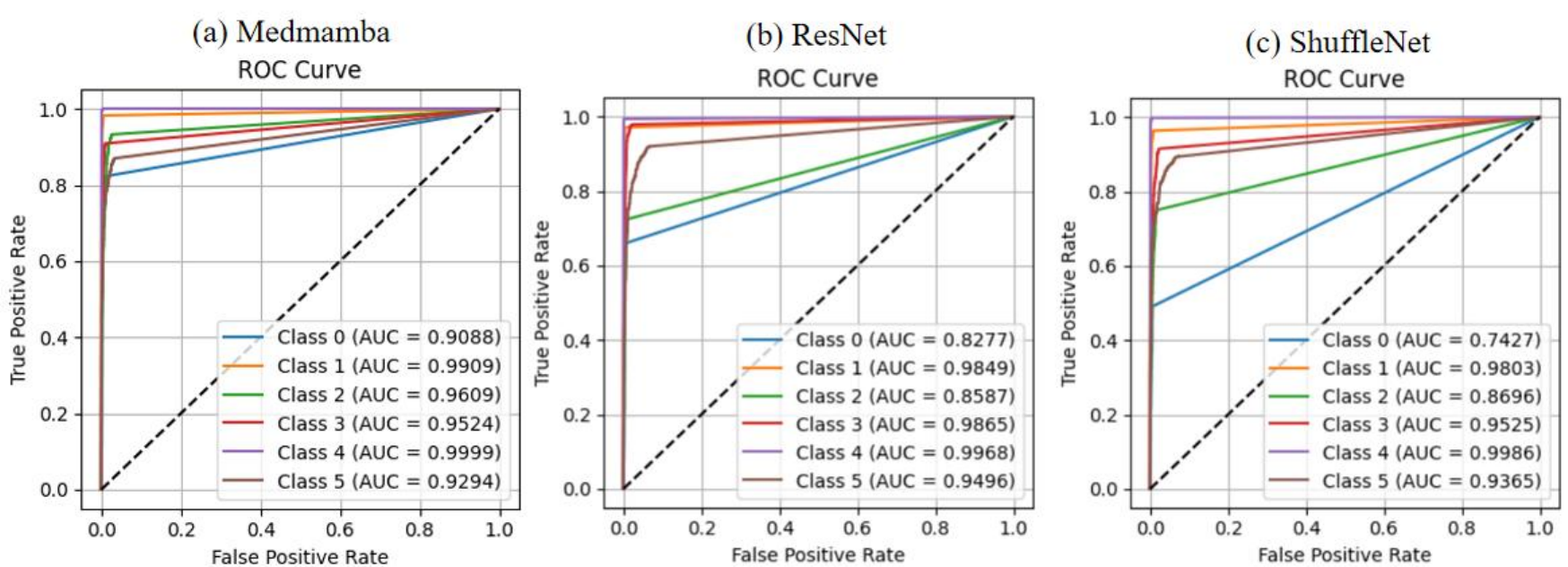}
    \includegraphics[width=0.9\linewidth]{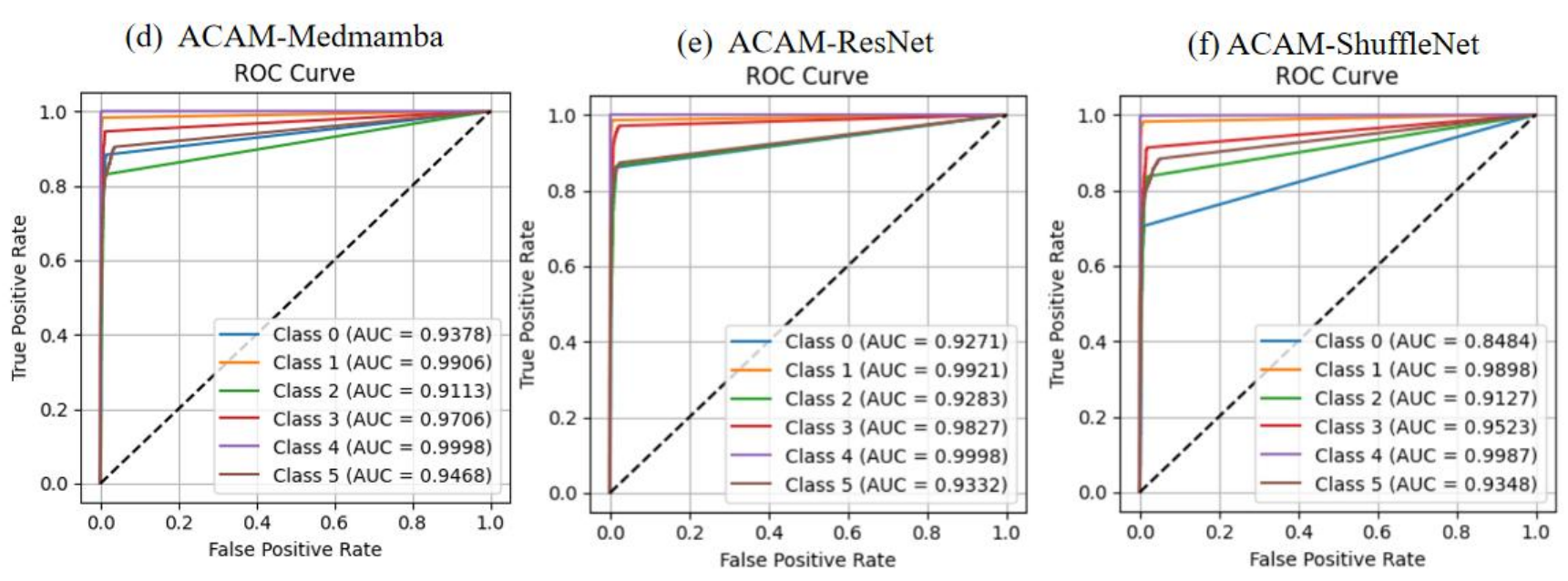}
    \caption{Comparison of ROC curves for three models before and after integrating the proposed module. (a)–(c) depict the classification performance of the baseline models, while (d)–(f) demonstrate the improvements achieved after incorporating the module.}
    \label{roc}
\end{figure}

\begin{figure}
    \centering
    \includegraphics[width=0.9\linewidth]{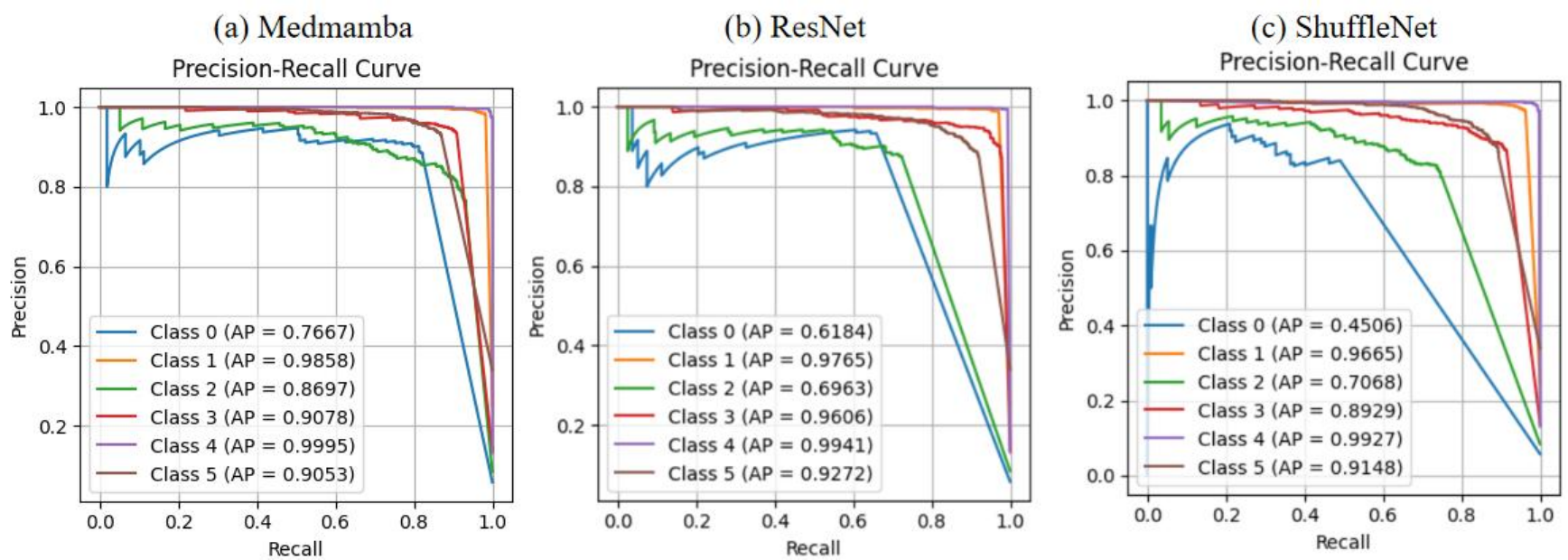}
    \includegraphics[width=0.9\linewidth]{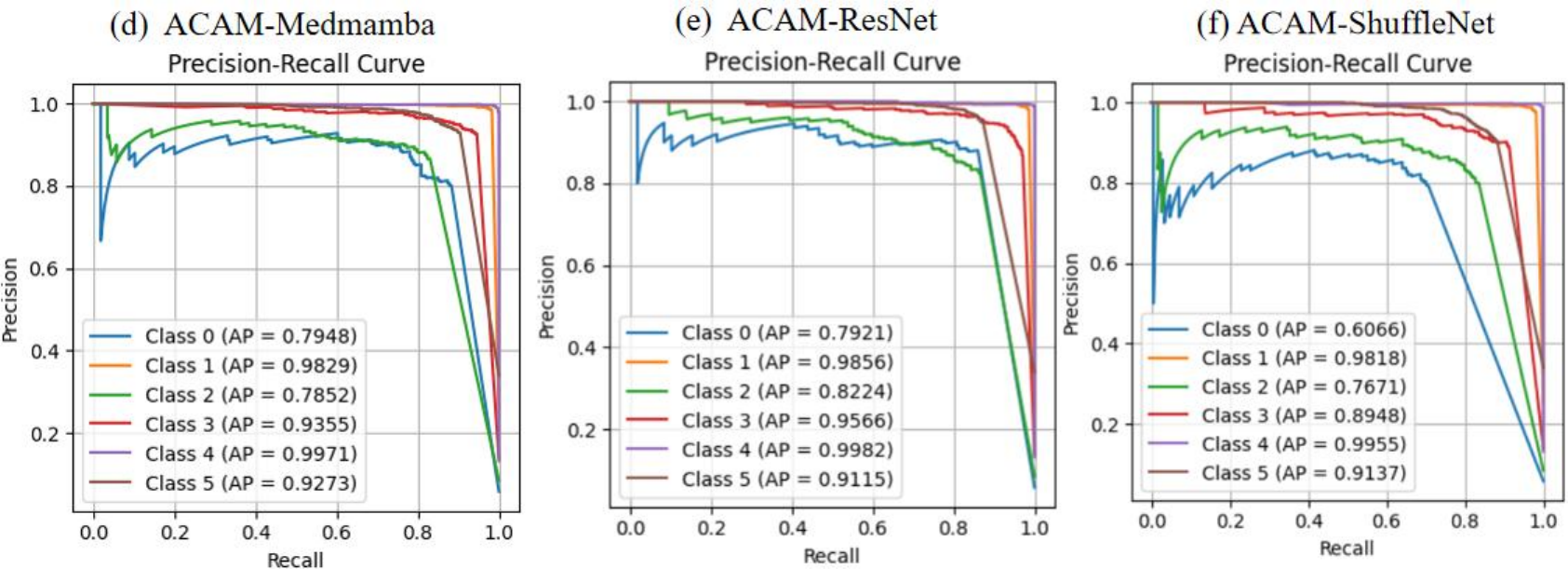}
    \caption{Comparison of PR curves for three models before and after integrating the proposed module. (a)–(c) illustrate the classification performance of the baseline models, while (d)–(f) demonstrate the improvements obtained after incorporating the module.}
    \label{pr}
\end{figure}

Analysis of the ROC curves, illustrated in Figure \ref{roc}, indicates that introducing the ACAM module leads to notable improvements in classification performance across all models. In the lightweight ShuffleNet, traditional ResNet, and advanced MedMamba, the trade-off between true positive rate (TPR) and false positive rate (FPR) is clearly improved for most classes. Specifically, integration with MedMamba significantly increases the AUC for classes 0, 3, and 5. ResNet shows clear performance gains for classes 0, 2, and 3, while ShuffleNet exhibits noticeable improvements for classes 0 and 2. These observations further confirm that ACAM provides consistent performance gains and robustness across different model architectures.
The precision–recall (PR) curves, presented in Figure \ref{pr}, demonstrate that the module substantially enhances classification performance for key classes. In ShuffleNet, ACAM effectively improves the precision-recall balance for classes 0, 1, and 2. For ResNet, notable improvements are observed in classes 0, 1, and 2, whereas in MedMamba, classes 0, 3, and 5 benefit significantly from the module. These results suggest that ACAM adaptively enhances the recognition of challenging samples according to the characteristics of different backbone networks, achieving higher recall while maintaining high precision, and thus demonstrating its general applicability and effectiveness in improving classification performance.

\subsection{Heatmap-Based Visualization and Analysis of Detailed Classification Results}
\begin{figure}
    \centering
    \includegraphics[width=0.7\linewidth]{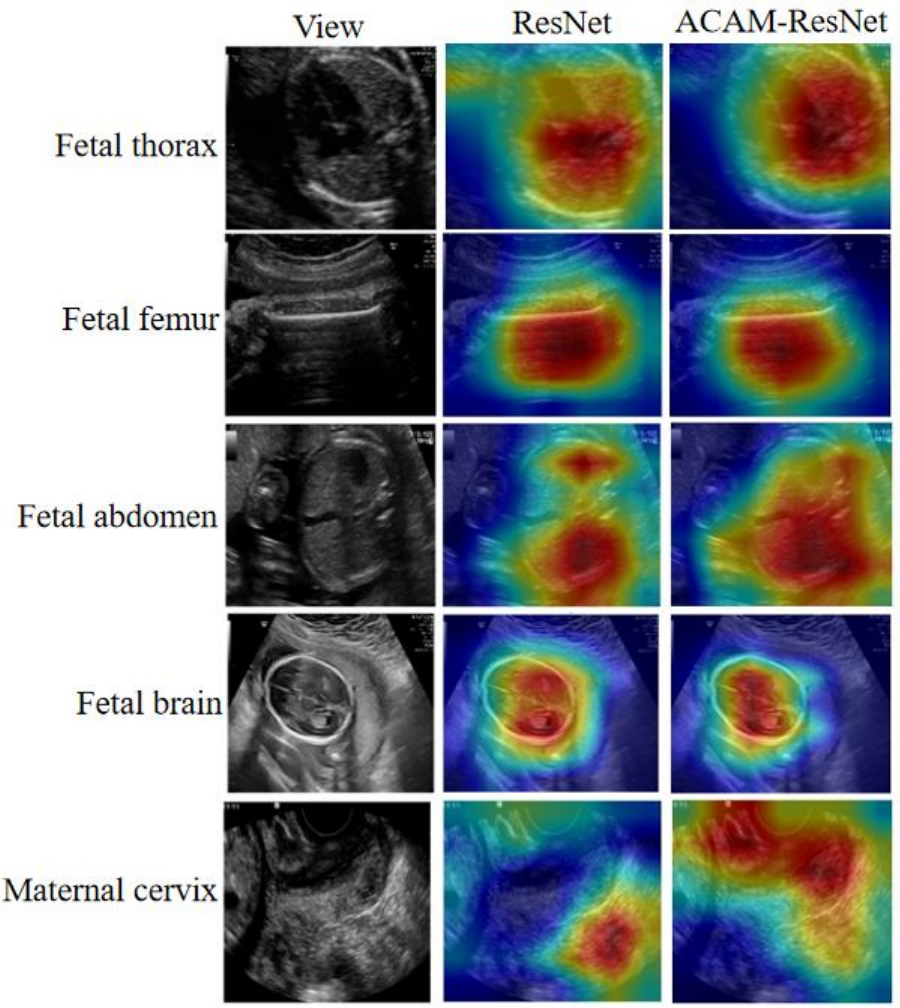}
    \caption{Heatmap visualizations of the ResNet model before and after integrating our module, illustrated on five representative image categories.}
    \label{fig5}
\end{figure}

To further validate the effectiveness of the proposed ACAM module, we employed the Grad-CAM method to visualize the model's attention regions. It is important to note that we primarily present visualization results based on ResNet, as the design of Grad-CAM relies on the spatial feature maps of convolutional layers, which enables the generation of heatmaps with better spatial correspondence and interpretability in convolutional models. As shown in Figure \ref{fig5}, the first column presents the original ultrasound images, while the second and third columns show the heatmaps generated by the baseline ResNet and the ACAM-enhanced ResNet (ACAM-ResNet), respectively. Experimental results indicate that, compared with the baseline ResNet whose attention regions are often dispersed or deviate from the target anatomical structures, ACAM-ResNet can more accurately focus on key regions relevant to clinical diagnosis.
In fetal thoracic planes, the heatmaps of the conventional ResNet tend to cover the entire thoracic area, whereas ACAM-ResNet significantly enhances attention to critical structures such as the heart and lungs. For fetal femoral planes, the baseline model may distribute attention to surrounding soft tissues, while the improved model accurately localizes the femoral shaft. In abdominal plane analysis, ACAM-ResNet demonstrates more explicit attention to the stomach bubble and umbilical cord insertion site, whereas the heatmaps of the conventional model are often blurred. For fetal brain planes, the improved model clearly focuses on the lateral ventricles and midline structures, avoiding interference from irrelevant brain tissues. Additionally, in maternal cervical planes, ACAM-ResNet effectively highlights the internal cervical os and the lumen structures, while the baseline model is easily distracted by surrounding tissues.

These visualization results fully demonstrate the efficacy of the ACAM module in guiding the model's attention, providing strong support for improving the accuracy and reliability of fetal ultrasound plane recognition. Table \ref{tab3} presents detailed classification performance across different planes. The model achieves high performance on most standard planes, with the best results observed in fetal femur (F1 = 0.9764) and fetal brain (F1 = 0.9869). In contrast, the recall for the cervical category (0.7442) is relatively low, indicating that discrimination in this plane remains challenging. Overall, the results suggest that the model exhibits strong robustness and effectiveness in handling diverse fetal ultrasound planes.

\begin{table}[!ht]
    \centering
    \caption{Ablation study results of our module integrated into different models, as well as comparisons with other models. The best-performing values are highlighted in bold}
    \begin{tabular}{p{4cm}p{3cm}p{3cm}p{3cm}}
    \hline
        Class & Recall & Precision & F1-score  \\ \hline
        Fetal abdomen & 0.8756 & 0.8224 & 0.8482  \\ 
        Fetal femur & 0.9702 & 0.9828 & 0.9764  \\ 
        Maternal cervix & 0.7442 & 0.9327 & 0.8279  \\ 
        Fetal thorax & 0.9343 & 0.9089 & 0.9214  \\ 
        Fetal brain & 0.9741 & 1.0000 & 0.9869  \\ 
        Other & 0.9267 & 0.8703 & 0.8976 \\ \hline
    \end{tabular}
    \label{tab3}
\end{table}

\section{Discussion}
\subsection{Module Significance}
In existing research on fetal ultrasound standard plane recognition, the majority of approaches rely on convolutional neural networks (CNNs) to exploit their advantages in texture and edge feature extraction \cite{venkatareddy2024explainable,diniz2020deep, wang2021deep}. However, these models typically assume that the input images exhibit stable quality and moderate contrast. In real-world clinical scenarios, factors such as fetal position, gestational age, device settings, and operator habits often result in considerable variations in image contrast, thereby obscuring or attenuating critical anatomical details. Clinicians frequently address this challenge by adjusting image contrast to highlight essential anatomical structures, thus obtaining clearer and more discriminative ultrasound images. This well-established clinical practice directly motivated the design of our ACAM.
Conventional data augmentation strategies may partially alleviate the adverse effects of contrast variation. Nevertheless, these approaches are primarily based on random transformations, representing a ``blind'' expansion that lacks adaptive correlation with image content. Consequently, they often fail to effectively capture clinically relevant anatomical details. ACAM is specifically designed to overcome this limitation. Its key contribution lies not only in enhancing image texture details but also in transforming contrast adjustment from a ``fixed preprocessing step'' into a ``content-aware dynamic decision.'' This paradigm shift enables the model to actively explore and integrate multiple potential contrast perspectives, thereby preserving strong discriminative capability even in challenging scenarios characterized by blurred structures, ambiguous boundaries, or low signal-to-noise ratios.

From a broader perspective, ACAM represents more than a technical enhancement for fetal plane classification; it embodies a modeling paradigm tailored to the intrinsic characteristics of medical imaging. By introducing dynamic modeling of contrast---a low-level imaging attribute---our approach brings deep learning models closer to the actual clinical process of image acquisition and interpretation, thereby providing novel insights for the advancement of medical artificial intelligence.

\subsection{Secondary Training Strategy}
Our model further supports an extended application. Specifically, records of clinicians’ contrast adjustment behaviors for each fetal ultrasound plane can be collected and subsequently employed to supervise the training of the convolutional module in the contrast generation stage (as illustrated in Fig.~1). During the subsequent classification phase, the parameters of the first-stage convolutional layers can be frozen. This secondary training strategy not only improves model performance but also enhances interpretability, as the feature generation process explicitly reflects clinicians’ operational preferences. Moreover, the approach demonstrates strong scalability, enabling its adaptation to data acquired from different devices or operators, thereby further strengthening the robustness and clinical applicability of the model.

\subsection{Limitations and Future Directions}
Although our method is capable of automatically generating multiple contrast values based on input images---thus enhancing the model’s sensitivity to fine-grained details---the number of generated contrast values is currently fixed. This design may constrain the adaptability of the model when exposed to extreme or previously unseen contrast variations. Future research could investigate more flexible contrast generation mechanisms, such as variable-size or continuously parameterized approaches, to better accommodate a wider spectrum of contrast distributions, thereby further improving model robustness and generalization capacity. Additionally, incorporating clinician adjustment records or prior clinical knowledge represents a promising direction to further enhance both interpretability and clinical relevance of the proposed method.

\section{Conclusion}
This work presents ACAM, a novel paradigm for fetal ultrasound plane classification that fundamentally mitigates performance degradation caused by low-contrast tissue boundaries. Inspired by clinical practice, where sonographers routinely adjust image contrast to obtain clearer and more discriminative views, we incorporate this insight into the design of ACAM. By integrating contrast adjustment directly into feature learning through a dynamically parameterized module, ACAM generates anatomically meaningful multi-contrast views guided by local texture cues, significantly enhancing detail discriminability without compromising semantic extraction. Its seamless integration across convolutional, lightweight, and modern architectures demonstrates universal effectiveness, with an average accuracy gain of 1.48\% validated on multi-center clinical data. Crucially, ACAM establishes clinical interpretability through two mechanisms: sigmoid-mapped parameters reflecting real sonographer adjustment ranges (1–3), and Grad-CAM visualizations confirming focused attention on diagnostic landmarks such as cardiac structures and femur shafts. Future work will explore physician-guided training via adjustment records and dynamic parameterization for broader contrast scenarios. ACAM provides a practical way of embedding imaging physics into deep learning pipelines, contributing to more reliable medical image analysis under heterogeneous clinical conditions.

\bibliographystyle{Frontiers-Harvard}
\bibliography{test}

\end{document}